\documentclass[runningheads]{llncs}
\usepackage{graphicx}
\usepackage{xspace}

\usepackage{tikz}
\usepackage{comment}
\usepackage{amsmath,amssymb} 
\usepackage{siunitx}
\usepackage{color}
\usepackage{xcolor}
\usepackage{subcaption}

\usepackage[flushleft]{threeparttable}
\usepackage{diagbox}
\usepackage{multirow}
\usepackage{tabulary}
\newcolumntype{x}[1]{>{\centering\arraybackslash}p{#1pt}}

\usepackage[hang,flushmargin]{footmisc}

\usepackage[accsupp]{axessibility}  

\usepackage[width=122mm,left=12mm,paperwidth=146mm,height=193mm,top=12mm,paperheight=217mm]{geometry}

\newcolumntype{L}[1]{>{\raggedright\let\newline\\\arraybackslash\hspace{0pt}}m{#1}}
\newcolumntype{C}[1]{>{\centering\let\newline\\\arraybackslash\hspace{0pt}}m{#1}}
\newcolumntype{R}[1]{>{\raggedleft\let\newline\\\arraybackslash\hspace{0pt}}m{#1}}

\newcommand{\app}{\raise.17ex\hbox{$\scriptstyle\sim$}}
\newcommand*{\x}{\mathsf{x}\mskip1mu}

\newlength\savewidth\newcommand\shline{\noalign{\global\savewidth\arrayrulewidth
  \global\arrayrulewidth 1pt}\hline\noalign{\global\arrayrulewidth\savewidth}}
\newcommand{\tablestyle}[2]{\setlength{\tabcolsep}{#1}\renewcommand{\arraystretch}{#2}\centering\footnotesize}

\newcommand{\eqn}[1]{Eq.~(\ref{#1})}
\newcommand{\fig}[1]{Fig.~\ref{#1}}
\newcommand{\sect}[1]{Sec.~\ref{#1}}
\newcommand{\tbl}[1]{Table~\ref{#1}}

\definecolor{darkgreen}{RGB}{114,210,115}
\definecolor{lime}{RGB}{34,139,34}
\definecolor{gucolor}{RGB}{50,200,0}

\makeatletter
\DeclareRobustCommand\onedot{\futurelet\@let@token\@onedot}
\def\@onedot{\ifx\@let@token.\else.\null\fi\xspace}
\def\eg{\emph{e.g}\onedot} 
\def\ie{\emph{i.e}\onedot} 
\def\cf{\emph{c.f}\onedot} 
\def\etc{\emph{etc}\onedot} 
\def\vs{\emph{vs}\onedot}
\def\wrt{w.r.t\onedot}

\def\etal{\emph{et al}\onedot}
\makeatother

\newcommand{\loss}[0]{\mathcal{L}}
\newcommand{\CM}[1]{\SI{#1}{\cm}}
\newcommand{\rete}[0]{$\ang{5}\,\CM{5}$}
\newcommand{\iou}[1]{$\text{IoU}_\text{#1}$}

\newcommand{\shp}[0]{\mathcal{P}}
\newcommand{\pcl}[0]{\mathcal{O}}

\newcommand{\rot}[0]{\mathbf{R}}
\newcommand{\trans}[0]{\mathbf{t}}
\newcommand{\size}[0]{\mathbf{s}}

\DeclareMathOperator{\avg}{avg}
\DeclareMathOperator{\Tr}{Tr}  

\definecolor{citecolor}{RGB}{34,139,34}  
\usepackage{orcidlink}

\begin{document}
\pagestyle{headings}
\mainmatter

\title{CATRE: Iterative Point Clouds Alignment for Category-level Object Pose Refinement}

\titlerunning{CATRE: Category-level Object Pose Refinement}
%
\author{Xingyu Liu$^{1,*}$\orcidlink{0000-0003-1156-2263} \quad Gu Wang$^{2,*}$\orcidlink{0000-0002-0759-0782} \quad Yi Li$^3$\orcidlink{0000-0002-7547-5073} \quad Xiangyang Ji$^1$\orcidlink{0000-0002-7333-9975}}
\institute{$^1$Tsinghua University, BNRist \quad $^2$JD.com \quad $^3$University of Washington\\
{\tt\small liuxy21@mails.tsinghua.edu.cn, guwang12@gmail.com, \\ 
yili.matrix@gmail.com, xyji@tsinghua.edu.cn}}
\authorrunning{X. Liu et al.}
%
\maketitle

\begin{abstract}
{\let\thefootnote\relax\footnotetext{$^{*}$ Equal contribution.}}
While category-level 9DoF object pose estimation has emerged recently, 
previous correspondence-based or direct regression methods are both limited in accuracy due to the huge intra-category variances in object shape and color, \etc.
Orthogonal to them, this work presents a category-level object pose and size refiner CATRE, which is able to iteratively enhance pose estimate from point clouds to produce accurate results.
Given an initial pose estimate, CATRE predicts a relative transformation between the initial pose and ground truth by means of aligning the partially observed point cloud and an abstract shape prior. 
In specific, we propose a novel disentangled architecture being aware of the inherent distinctions between rotation and translation/size estimation.
Extensive experiments show that our approach remarkably outperforms state-of-the-art methods on REAL275, CAMERA25, and LM benchmarks up to a speed of $\approx$85.32Hz, and achieves competitive results on category-level tracking.
We further demonstrate that CATRE can perform pose refinement on unseen category.
Code and trained models are available.\footnote{ \href{https://github.com/THU-DA-6D-Pose-Group/CATRE.git}{https://github.com/THU-DA-6D-Pose-Group/CATRE.git}}
\end{abstract}
\section{Introduction}

Estimating the 6DoF pose, \ie, 3DoF orientation and 3DoF localization, of an object in Euclidean space plays a vital role in robotic manipulation~\cite{tremblay2018deep,du2021vision}, 3D scene understanding~\cite{nie2020total3dunderstanding,huang2018cooperative} and augmented/virtual reality~\cite{marchand2015pose,su2019arvr}.
The vast majority of previous works~\cite{li20deepim_ijcv,zakharov2019dpod,hybridpose,posecnn,Wang_2021_self6dpp,Wang_2021_GDRN,li2019cdpn,wang2019densefusion,peng2019pvnet} study with instance-level object pose estimation, which can be decomposed by two procedures: initial pose estimation and pose refinement.
Despite considerable progress has been made in instance-level settings, the generalizability and scalability \wrt unseen objects are limited due to the high dependency of known exact CAD models during both training and inference.

To alleviate this problem,
recently, increasing attention has been received on category-level 9DoF pose (\ie, 6DoF pose and 3DoF size) estimation which aims to handle novel instances among known categories without requiring CAD models for test. 
As an early proposed work, Wang \etal~\cite{wang2019nocs} predict the normalized object coordinates (NOCS) in image space and then solve the pose and size by matching NOCS against observed depth with Umeyama algorithm~\cite{umeyama}.
Afterwards,
several works \cite{Tian_ECCV20_DeformNet,chen_ICCV21_sgpa,Wang_IROS21_Cascaded,fan2021acr} attempt to deform the shape prior (\ie, the mean shape) of a category towards observed instances to improve the accuracy of correspondences matching.
However, those methods highly rely on the RANSAC procedure to remove outliers thus making them non-differentiable and time-consuming.
Contrary to correspondence-based methods, some more recent works~\cite{Chen_CVPR21_FSNet,Lin_ICCV21_DualPoseNet} propose to directly regress pose and size to achieve a higher speed during inference.
Nevertheless, these end-to-end approaches are oftentimes error-prone in that they are not sufficiently aware of the inherent distinctions between rotation and translation/size estimation.
To summarize, compared with the milestones achieved by state of the art in instance-level works, category-level pose estimation methods are still limited in accuracy.

Previously, when the CAD model is available, one common way of enhancing pose accuracy is to apply a post-refinement step through matching the rendered results against observed images given initial estimates, 
which has been widely explored in both traditional~\cite{ICP_tpami92,segal2009generalized} and learning-based~\cite{li20deepim_ijcv,Wen_IROS20_se3tracknet,labbe2020cosypose,iwase2021repose} methods.
Being motivated by this, we seek to tackle the above problem by investigating object pose refinement at the category level.
However, traditional object pose refinement methods rely on the CAD model to perform render-and-compare, which is not accessible when we conduct category-level object pose estimation.

\begin{figure}[tb!]
\begin{center}
  \includegraphics[width=0.95\linewidth]{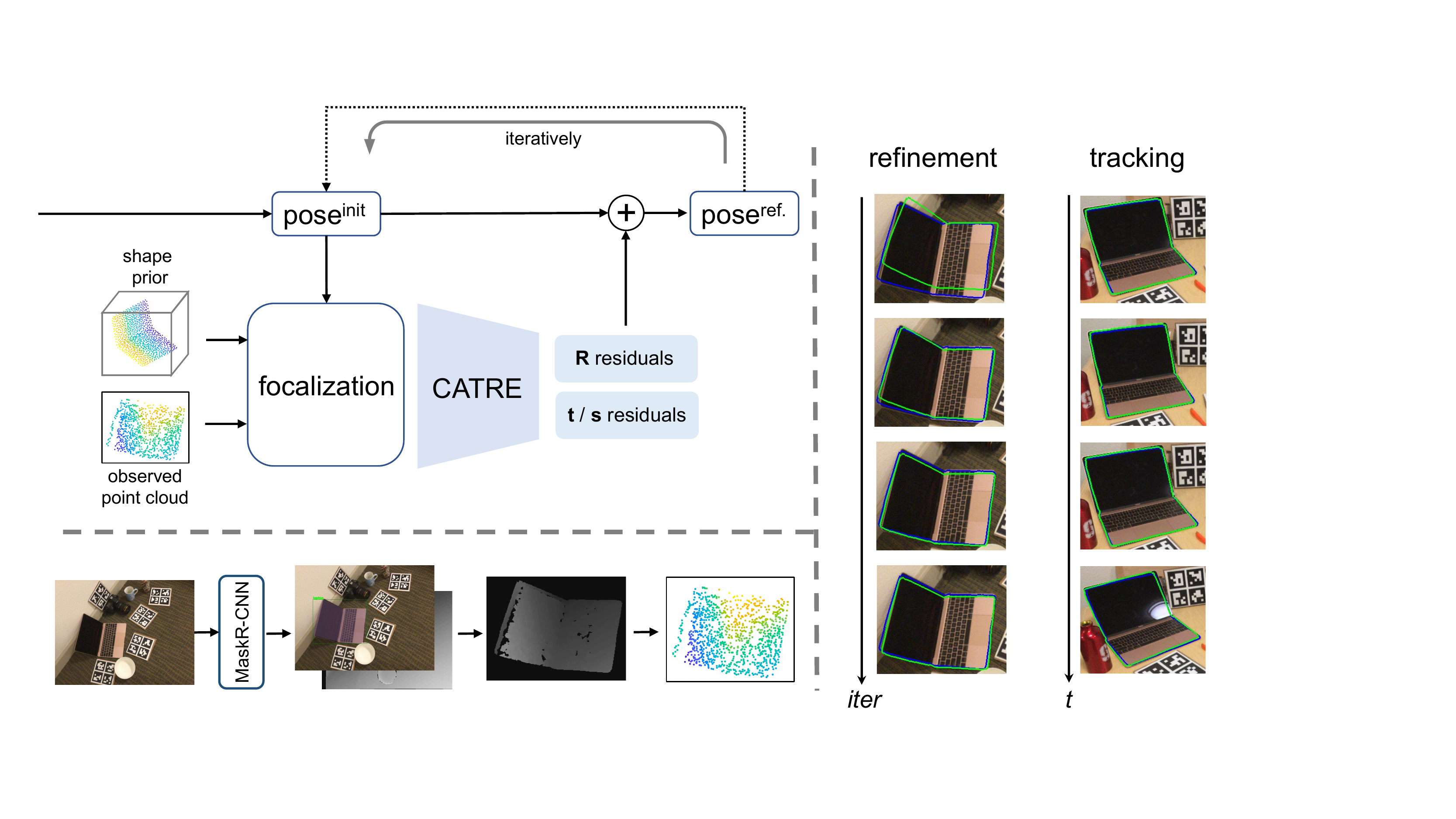}
\end{center}
  \caption{
  The framework of CATRE.
  Given an initial pose and size estimate [$\rot_{init}|\trans_{init}|\size_{init}$], CATRE predicts a relative transformation [$\rot_{\Delta}|\trans_{\Delta}|\size_{\Delta}$] by iteratively aligning two focalized point clouds, \ie, shape prior and observed point cloud. The framework can perform pose refinement and tracking, where blue and green contours reflect ground-truth and predicted poses, respectively.}
\label{fig:pipeline}
\end{figure}

To solve this dilemma,
we propose a novel method for \emph{CAT}egory-level object pose \emph{RE}finement 
(\emph{CATRE}), leveraging the abstract shape prior information instead of exact CAD models.
As shown in \fig{fig:pipeline}, 
we aim to directly regress the relative pose and size transformations by aligning the partially observed point cloud against the transformed shape prior given initial pose and size predictions. 
In specific, we first use the initial pose prediction to focalize the shape prior and observed point cloud into a 
limited range.
Then, a Siamese PointNet-based~\cite{qi2017pointnet} encoder is employed to map the two input point clouds into a common feature space while maintaining relevant features for rigid transformation.
Finally, we design two distinct heads to predict relative rotation and translation/size transformations in a disentangled manner.
This is based on the observation that rotation is heavily reliant on point-level local geometry whereas translation and size reside with the object-level global feature.
Besides, the procedure of refinement is conducted iteratively to achieve more accurate results. 

Extensive experiments demonstrate that our proposed method can accurately yet efficiently refine and track category-level object poses.
Exemplarily, we achieve a significant improvement over the baseline SPD~\cite{Tian_ECCV20_DeformNet} with an improvement of 26.8\% on the strict $\ang{5}\,\CM{2}$ metric and 14.5\% on the \iou{75} metric on REAL275 dataset.

To sum up, our contributions are threefold:
{i) }To the best of our knowledge, we propose the first versatile pipeline for category-level pose refinement based on point clouds leveraging abstract shape prior, 
without requiring exact CAD models during training or inference.
{ii) }For the learning of relative transformation, a 
pose-guided focalization strategy is proposed to calibrate input point clouds. 
We further introduce a novel disentangled architecture being aware of the inherent attributions in pose estimation.
Thanks to these key ingredients, we overcome the drawback of being error-prone while maintaining the high speed of direct regression.
{iii) }Our versatile framework can also perform category-level pose tracking and achieve competitive performance against state of the art, whilst at $7\x$ faster speed. 
Meanwhile, CATRE can be naturally extended to instance-level and unseen category pose refinement leveraging appropriate shape priors.

\section{Related Work}

\paragraph{Category-level Object Pose Estimation}
Category-level pose estimation aims to predict the 9DoF pose of a novel instance without the aid of its CAD model. 
Existing works can be generally categorized to correspondence-based~\cite{wang2019nocs,chen_ICCV21_sgpa,Tian_ECCV20_DeformNet,Wang_IROS21_Cascaded} and direct regression~\cite{Lin_ICCV21_DualPoseNet,Chen_CVPR21_FSNet,Chen_CVPR20_CASS} approaches. 
Correspondence-based approaches first predict dense object coordinates in a normalized canonical space (NOCS)~\cite{wang2019nocs} and then solve the pose by Umeyama algorithm~\cite{umeyama}. 
Notably, SPD~\cite{Tian_ECCV20_DeformNet} proposes to extract a categorical shape prior and adapt it to various instances via deformation prediction, in an effort to improve the matching of correspondences. 
On the other hand, direct regression methods predict object pose in an end-to-end manner, 
achieving a higher inference speed.
For instance, 
FSNet~\cite{Chen_CVPR21_FSNet} decouples rotation prediction into two orthogonal axes estimation.
DualPoseNet~\cite{Lin_ICCV21_DualPoseNet} exploits spherical convolutions to explicitly regress the pose with an auxiliary task of canonicalizing the observed point cloud meanwhile.

While this field has emerged recently, the accuracy of 
estimating category-level pose is still far from satisfactory compared with instance-level settings.
Orthogonally, this work builds an end-to-end category-level pose refinement pipeline which can largely enhance the performance whilst being fast.

\paragraph{Object Pose Refinement}
In instance-level cases~\cite{li20deepim_ijcv,labbe2020cosypose,Wen_IROS20_se3tracknet,zakharov2019dpod,hybridpose,posecnn,Wang_2021_self6dpp}, object pose refinement has demonstrated to be very effective.
Notably, DeepIM~\cite{li20deepim_ijcv} predicts pose transformation by comparing the rendered image against the observed image. 
se(3)-TrackNet~\cite{Wen_IROS20_se3tracknet} further leverages this render-and-compare strategy to object tracking using RGB-D images.
Extended from DeepIM, CosyPose~\cite{labbe2020cosypose} conquers the leaderboard of BOP Challenge 2020~\cite{hodan2020bop}, showing the powerful capability of the render-and-compare technique.

Different from the instance-level settings, category-level pose refinement is still rarely explored.
DualPoseNet~\cite{Lin_ICCV21_DualPoseNet} refines their prediction by a self-adaptive pose consistency loss, which is only applicable to specific networks.
However, our work is generalizable to various kinds of baselines~\cite{wang2019nocs,Tian_ECCV20_DeformNet,Lin_ICCV21_DualPoseNet} without re-training.

\paragraph{Category-level Object Pose Tracking}
6-PACK~\cite{Wang_ICRA20_6PACK} performs category-level tracking by anchor-based keypoints generation and matching between adjacent frames.
CAPTRA~\cite{weng2021captra} tracks rigid and articulated objects by employing two separate networks to predict the interframe rotation and normalized coordinates respectively.
Recently, iCaps~\cite{deng2022icaps} leverages a particle ﬁltering framework to estimate and track category-level pose and size.
Note that CAPTRA could be adapted for refinement in theory, although there are no relevant experiments. 
Still, there might be some limitations. 
First, it relies on CAD models to provide supervision signals (\ie, NOCS map) for training, while CATRE does not need exact CAD models during training or inference. 
Moreover, it was proposed for tracking where the errors between adjacent frames are minor, so it is unclear if CAPTRA could handle noises from other estimation methods. 
However, as CATRE directly regresses the relative pose, it can perform iterative inference with a faster speed and more robust performance \wrt noises.

\paragraph{Point Cloud Registration}
A closely related field to depth-based refinement is point cloud registration. 
Traditional methods like ICP~\cite{ICP_tpami92} and its variants~\cite{bouaziz2013sparse,segal2009generalized,rusinkiewicz2001efficient} require reliable pose initialization.
To overcome this problem, learning-based methods are proposed recently~\cite{Wang_iccv19_DCP,Aoki_cvpr19_PointNetLK,Wang_NeurIPS19_PRNet,Choy_CVPR20_DGR}.

Works like \cite{Groueix_2018_ECCV,NEURIPS2021_shape_registration} use the non-overlapped prior as CATRE does, 
but they usually learn the deformation rather than the relative pose.
Besides, shape prior in CATRE can be very different from the objects, varying from instance-specific keypoints to categorical mean shape to generic bounding box corners.

\section{Category-level Object Pose Refinement}
\begin{figure}[tb!]
\begin{center}
  \includegraphics[width=\linewidth]{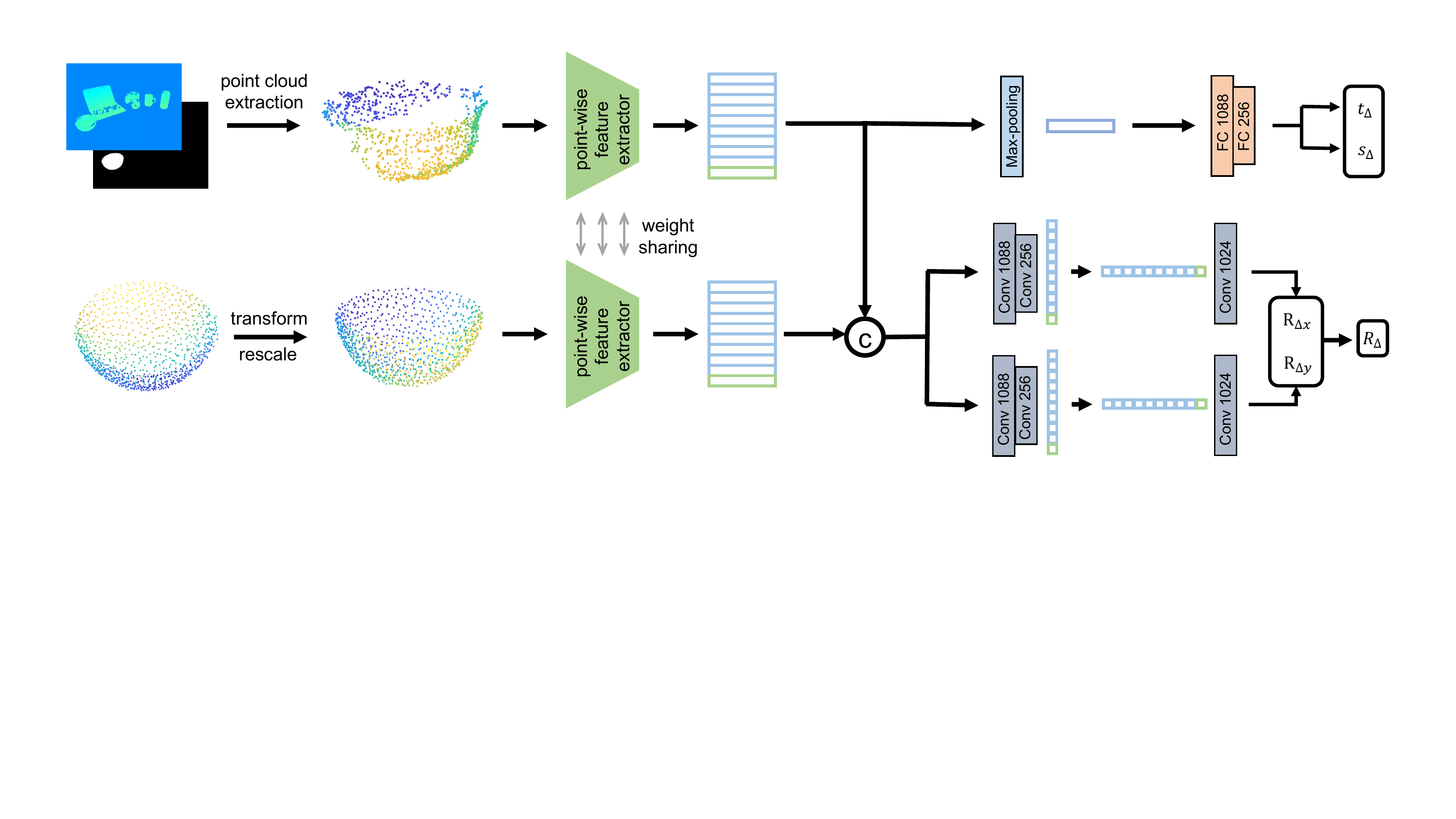}
\end{center}
  \caption{The network architecture of CATRE. Taking an observed point cloud $\pcl$ and shape prior $\shp$ as input, CATRE first focalizes them into a limited range using initial pose prediction [$\rot_{init}|\trans_{init}|\size_{init}$]. Then, we employ a Siamese geometric encoder followed by two disentangled heads to predict relative pose transformation [$\rot_{\Delta}|\trans_{\Delta}|\size_{\Delta}$].}
\label{fig:network}
\end{figure}

This section first addresses the problem formulation and framework overview (\fig{fig:pipeline}), and then describes the key ingredients of our approach (\fig{fig:network}) for category-level object pose refinement.
Afterwards, we present the training and testing protocol and show that CATRE can be naturally applied to pose tracking.

\subsection{Problem Formulation}
\label{sec:problem_form}
Given an initial pose and size estimate $[\rot_{init}|\trans_{init}|\size_{init}]$ and the observed point cloud $\mathcal{O}$ of an object, our goal is to predict a relative transformation $[\rot_{\Delta}|\trans_{\Delta}|\size_{\Delta}]$ between the initial prediction and ground truth $[\rot_{gt}|\trans_{gt}|\size_{gt}]$ leveraging a shape prior $\mathcal{P}$, as shown in \eqn{eq:catre}.
\begin{equation}
\label{eq:catre}
    [\rot_{\Delta}|\trans_{\Delta}|\size_{\Delta}] = CATRE([\rot_{init}|\trans_{init}|\size_{init}], \mathcal{O}, \mathcal{P}).
\end{equation}
Thereby, $\rot \in SO(3)$, $\trans \in \mathbb{R}^3$ and $\size \in \mathbb{R}^3$ respectively represents rotation, translation and size. 
The shape prior $\mathcal{P}$ can be the mean shape of a given category or a generic skeleton like bounding box corners. 
We use the categorical mean shape (\cf Sec. \ref{sec:overview}) as the prior information in our main experiments and illustrations.

\subsection{Overview of CATRE}
\label{sec:overview}
Recapping \fig{fig:pipeline}, we first employ an off-the-shelf instance segmentation network (\eg, Mask R-CNN~\cite{maskrcnn}) to cut the object of interest from the observed depth map. 
Then we back-project the depth into camera space and sample $N_o$ points within a ball near to the object following~\cite{g2lnet,weng2021captra}.
The center and radius of this ball are determined by the initial pose estimate.
To this end, we effectively remove outliers and obtain the observed point cloud $\mathcal{O}=\{o_i \in \mathbb{R}^3\}_{i=1}^{N_o}$.

To estimate the pose and size transformations based on the observed point cloud, we leverage the publicly available shape prior $\mathcal{P}=\{p_j \in \mathbb{R}^3\}_{j=1}^{N_p}$ from \cite{Tian_ECCV20_DeformNet}.
Thereby, a mean shape is reconstructed from the mean latent embedding for each category trained with a large amount of synthetic point cloud models from ShapeNet~\cite{chang2015shapenet} using an encoder-decoder framework.
We normalize the size of the shape prior such that the length of each side of its bounding box is 1.

Taking the observed point cloud $\mathcal{O}$, and the prior point cloud $\mathcal{P}$ as input, our CATRE network (\fig{fig:network}) essentially predicts the relative transformation to iteratively refine the initial estimate towards the target pose and size.

\subsection{Disentangled Transformation Learning}
\label{sec:key_ingredients}
\subsubsection{Point Clouds Focalization}
As the localization of input point clouds can vary dramatically in camera space, it would be very hard for the network to exploit useful information for learning relative transformation. 
Inspired by the image-space zooming in operation in DeepIM~\cite{li20deepim_ijcv}, we propose a novel point cloud focalization strategy to limit the range of input point clouds.
Specifically, after transforming the prior point cloud $\mathcal{P}$ into camera space with the initial pose and size, 
we simultaneously shift the observed and estimated point clouds using the initial translation.
Therefore, the input observed point cloud $\mathcal{\hat{O}}$ and prior point cloud $\mathcal{\hat{P}}$ can be obtained as follows
\begin{equation}
\begin{aligned}
\mathcal{\hat{O}} = & \{\hat{o}_i \mid \hat{o}_i = o_i - \trans_{init}\}_{i=1}^{N_o}, \\
\mathcal{\hat{P}} = & \{\hat{p}_j \mid \hat{p}_j=\size_{init} \odot \rot_{init}p_j\}_{j=1}^{N_p},
\end{aligned}
\end{equation}
where $\odot$ denotes element-wise product.
Geometrically, $\mathcal{\hat{P}}$ and $\mathcal{\hat{O}}$ contain all the information needed for learning the relative transformation between the initial estimate and the target pose, no matter how we shift them. 
By fixing the center of $\mathcal{\hat{P}}$ to the origin, we can let the network focalize on the limited range around the origin, thus reducing the difficulty of learning for the network.
Notice that, compared with similar strategies in image space proposed for instance-level 6D pose refinement and tracking~\cite{li20deepim_ijcv,Wen_IROS20_se3tracknet}, which rely on rendering with known CAD models and the costly operation of cropping and resizing, our point cloud focalization is much simpler yet more efficient.

\subsubsection{Shared Encoder}
Inspired by works on learning-based point cloud registration~\cite{lee2021deeppro,Wang_iccv19_DCP,sarode2019pcrnet}, we employ a shared encoder to respectively extract embeddings
$F_p \in \mathbb{R}^{N_p \times d}$ and $F_o \in \mathbb{R}^{N_o \times d}$ from  $\mathcal{\hat{P}}$ and $\mathcal{\hat{O}}$ for efficiently encoding the geometric information locally and globally.
Here $d=1088$ denotes the output dimension.
In point cloud registration, both point-based encoder~\cite{qi2017pointnet} and graph-based encoder~\cite{3DGC} have been widely used.
However, 
we experimentally find 3D-GCN~\cite{3DGC} performing poorly in our work, as it is locally sensitive whilst the local graph structures of the partially observed point cloud and the full shape prior share little similarity.
Hence we choose the simpler PointNet~\cite{qi2017pointnet} as the backbone
for better generalizability.
Concretely, PointNet maps each point from $\mathbb{R}^3$ into a high-dimensional space, and generates an additional global feature by aggregating features from all points. 
The global feature is repeated $N$ times and appended to each point-wise feature, where $N$ denotes the number of input points. 
For details of the employed encoder, please kindly refer to the supplementary material.

By the prominent weight sharing strategy, we map $\mathcal{\hat{P}}$ and $\mathcal{\hat{O}}$ into a common feature space while remaining sensitive to relevant features for rigid transformation.
Considering that rotation is different from translation and size in essence, we introduce a disentangled transformation learning strategy for them 
by means of 
two different heads namely \emph{Rot-Head} and \emph{TS-Head} in the following.

\subsubsection{Rotation Prediction}
As aforementioned, 
$\mathcal{\hat{O}}$ encodes the information of $\rot_{gt}$, and $\mathcal{\hat{P}}$ contains $\rot_{init}$, so the network can predict $\rot_{\Delta}$ by comparing $\mathcal{\hat{O}}$ and $\mathcal{\hat{P}}$.
First, we concatenate $F_o$ and $F_p$ along the point dimension to get a unified per-point feature
$F_{op} \in \mathbb{R}^{(N_o+N_p) \times d}$. 
Then, the feature is fed to Rot-Head to predict a continuous 6-dimensional rotation representation $[\mathbf{r}_{\Delta}^x|\mathbf{r}_{\Delta}^y]$ following~\cite{zhou2019continuity,Chen_CVPR21_FSNet}, \ie, the first two columns of the rotation matrix $\rot_\Delta$. 
Finally, the relative rotation matrix $\rot_\Delta$ is recovered as in \cite{zhou2019continuity}, 
and the refined rotation prediction can be computed by $\rot_{est}=\rot_{\Delta}\rot_{init}$, where $\rot_{est}$ is used for loss calculation and re-assigned as $\rot_{init}$ for the next iteration.

Noteworthy, since the prediction of relative rotation is sensitive to local geometry, it is crucial how the network fuses the information of $F_{op}$ at point level.
Instead of using non-parametric operations like max-pooling which loses the local information severely, we opt for a trainable way to aggregate the point-wise features in Rot-Head. 
In particular, we employ two parallel branches for Rot-Head to obtain axis-wise rotation predictions $\mathbf{r}_{\Delta}^x$ and $\mathbf{r}_{\Delta}^y$, respectively.
Each branch is comprised of two 1D convolutional layers each followed by Group Normalization~\cite{wu2018group} and GELU~\cite{hendrycks2016gaussian} activation to generate an implicit per-point 
rotation prediction $f_{op} \in \mathbb{R}^{(N_o+N_p) \times 3}$ first.
Then the explicit global axis-wise prediction is aggregated from the permuted point-wise implicit prediction $f_{op}^{'} \in \mathbb{R}^{3 \times (N_o+N_p)}$ using an extra 1D convolutional layer, reducing the dimension from $3 \times (N_o+N_p)$ to $3 \times 1$. 
In this way, the network can preserve the local information regarding the relative rotation between $\mathcal{\hat{P}}$ and $\mathcal{\hat{O}}$ in a trainable way.

\subsubsection{Translation and Size Prediction}
Since $\pcl$ is shifted to $\mathcal{\hat{O}}$ by $\trans_{init}$, 
the translation transformation $\trans_{\Delta}$ can be derived from $F_o$ directly. 
Different from predicting rotation by aggregating point-level information, 
we argue that translation and size should be treated differently in that they are both global attributions at the object level. 
Therefore, TS-Head first extracts a global feature $f_o$ from $F_o$ by performing max-pooling along the point dimension.
Then, the initial scale is explicitly appended to $f_o$ for predicting the size transformation $\size_{\Delta}$.
Afterwards, two fully connected (FC) layers are applied to the global feature, reducing the dimension from 1091 to 256.
Finally, two parallel FC layers output $\trans_{\Delta}$ and $\size_{\Delta}$ separately.
The predicted translation and size are obtained by $\trans_{est} = \trans_{init} + \trans_{\Delta}$ and $\size_{est} = \size_{init} + \size_{\Delta}$ accordingly.
Similar to the rotation prediction, they are also re-assigned as the initial estimates for the next iteration.

\subsection{Training and Testing Protocol}
\label{sec:train_test}
\subsubsection{Pose Loss} In works for direct pose regression, the design of loss function is crucial for optimization~\cite{Wang_2021_GDRN}.
Apart from using an angular distance based loss on rotation~\cite{huynh2009metrics}, and $L_1$ losses for translation and scale, we additionally leverage a novel loss modified based on point matching~\cite{li20deepim_ijcv} to couple the estimation of $[\rot|\trans|\size]$. 
To summarize, the overall loss function can be written as
\begin{equation}
    \mathcal{L} = \mathcal{L}_{pm} + \mathcal{L}_{\rot} + \mathcal{L}_{\trans} + \mathcal{L}_{\size},
\end{equation}
where
\begin{equation}
\begin{cases}
    \mathcal{L}_{pm} &= \underset{\textbf{x} \in \mathcal{P}}{\avg}\|(\size_{gt} \odot \rot_{gt}\textbf{x} + \trans_{gt}) - (\size_{est} \odot \rot_{est}\textbf{x} + \trans_{est})\|_1, \\ 
    \mathcal{L}_{\rot} &= \frac{3-\Tr(\rot_{gt}\rot_{est}^T)}{4}, \\
    \mathcal{L}_{\trans} &= \|\trans_{gt} - \trans_{est} \|_1, \\
    \mathcal{L}_{\size} &= \|\size_{gt} - \size_{est} \|_1.
\end{cases}
\end{equation}
Here $\odot$ denotes element-wise product.
For symmetric objects, we rotate $\rot_{gt}$ along the known symmetry axis to find the best match for $\rot_{est}$. 

\subsubsection{Iterative Training and Testing} 
We employ an iterative training and testing strategy for pose refinement inspired by~\cite{li2018deepim_ECCV},
namely the current prediction is
re-assigned as the initial pose for the next iteration.
By repeating this procedure multiple times, \ie, 4 in this work, 
not only the diversity of training pose error distribution is enriched,
but also more accurate and robust inference results can be obtained.
To decouple CATRE from different initial pose estimation methods,
during training we add random Gaussian noise to $[\rot_{gt}|\trans_{gt}|\size_{gt}]$ as initial pose in the first iteration.
The network can therefore refine the pose prediction from various methods~\cite{wang2019nocs,Lin_ICCV21_DualPoseNet,Tian_ECCV20_DeformNet} consistently without re-training (see also \tbl{tab:cat_initpose}).

\subsection{Pose Tracking}
\label{sec:tracking}
It is natural to apply CATRE to tracking due to the similarity of pose refinement and tracking.
Given a live point cloud stream and the 9DoF pose estimate $[\rot_{t}|\trans_{t}|\size_{t}]$ of an object at frame $t$, the target of category-level pose tracking is to estimate the object pose $[\rot_{t+1}|\trans_{t+1}|\size_{t+1}]$ of the next frame $t+1$.
Similar to pose refinement, we use $[\rot_{t}|\trans_{t}|\size_{t}]$ as the initial pose to focalize $\shp$ and $\pcl$, and predict the relative pose transformation between frame $t$ and $t+1$.

During inference, we jitter the ground-truth pose of the first frame with the same Gaussian noise distribution in \cite{weng2021captra}. 
The pose is re-initialized if the adjacent frames are not consecutive.
\section{Experiments}

\subsection{Experimental Setup}
\paragraph{Implementation Details}
All the experiments are implemented using PyTorch~\cite{paszke2019pytorch}.
We train our model using Ranger optimizer~\cite{ranger1,ranger2,ranger3} with a base learning rate of $1 \times 10^{-4}$, annealed at 72\% of the training epoch using a cosine schedule~\cite{loshchilov-ICLR17SGDR}.
The total training epoch is set to 120 with a batch size of 16.
We employ some basic strategies for depth augmentation, \eg, randomly dropping depth values, adding Gaussian noise, and randomly filling 0 points, as well as some pose augmentations like random scaling, random rotation and translation perturbation as in \cite{Chen_CVPR21_FSNet}.
Unless otherwise specified, the size of input images are 480$\times$640, $N_o$ and $N_p$ are empirically set to 1024, and a single model is trained for all categories.

\paragraph{Datasets}
We conduct experiments on 4 datasets: REAL275~\cite{wang2019nocs}, CAMERA25~\cite{wang2019nocs}, LM~\cite{hinterstoisser2012}, and YCB-V~\cite{posecnn}.
\emph{REAL275} is a widely-used real-world category-level pose benchmark containing 4.3k images of 7 scenes for training, and 2.75k images of 6 scenes for testing. 
\emph{CAMERA25} is a synthetic dataset generated by a context-aware mixed reality approach. 
There are 300k composite images of 1085 object instances in total, where 25k images and 184 instances are used for evaluation.
REAL275 and CAMERA25 share the same object categories, \ie, bottle, bowl, camera, can, laptop, and mug. 
\emph{LM} dataset consists of 13 object instances, each of which is labeled around the center of a sequence of  $\approx$1.2k cluttered images.
We follow~\cite{brachmann2016uncertainty} to split $\approx$15\% images for training and $\approx$85\% images for testing.
Additionally, 1k rendered images for each instance are used to assist training as in~\cite{li2019cdpn}.
\emph{YCB-V} dataset is very challenging due to severe occlusions and various lighting conditions. 
We select two objects, \ie, master chef can and cracker box, out of 21 instances for unseen category refinement.

\paragraph{Metrics}
On REAL275 and CAMERA25, we follow \cite{wang2019nocs} to report the mean Average Precision (mAP) of intersection over union (IoU) metric at different thresholds\footnote{Note that there is a small mistake in the original IoU evaluation code of \cite{wang2019nocs}, we recalculated the IoU metrics as in \cite{peng2022self}.}, as well as mAP at $n^\circ\,m\,\CM{}$ for evaluating pose accuracy.
Additionally, for the task of tracking, we report mIoU, $\rot_{err}$, and $\trans_{err}$ as in~\cite{weng2021captra}, respectively standing for the mean IoU of ground-truth and predicted bounding boxes, average rotation error, and average translation error.
On LM, the ADD metric~\cite{hinterstoisser2012} is used to measure whether the average offset of transformed model points is less than 10\% of the object diameter.
For symmetric objects, the ADD-S metric~\cite{hodan2016evaluation} is employed.
On YCB-V, we use the area under the accuracy curves (AUC) of ADD(-S) metric varying the threshold from 0 to \CM{10} as in \cite{posecnn}.

\subsection{Category-level Pose Refinement}
\label{sec:cat_ref}
\subsubsection{Results}

\begin{table*}[t!]
\centering
\caption[caption]{Results of category-level pose refinement on REAL275 and CAMERA25. SPD$^*$ denotes our  re-implementation of SPD~\cite{Tian_ECCV20_DeformNet}.}
\label{tab:cat_ref_sota}
\scalebox{0.95}{
\tablestyle{3pt}{1.0}
\begin{tabular}{@{}l|l|c|c|cccc@{}}
    \shline
    Dataset & Method & Refine &  \iou{75} & $\ang{5}\,\CM{2}$ & $\ang{5}\,\CM{5}$ & $\ang{10}\,\CM{2}$ & $\ang{10}\,\CM{5}$ \\
    \hline
    \multirow{7}{*}{REAL275} & NOCS~\cite{wang2019nocs}&  & 9.4 & 7.2 & 10.0 & 13.8 & 25.2 \\
    & DualPoseNet~\cite{Lin_ICCV21_DualPoseNet} & \checkmark & 30.8 & 29.3 & 35.9 & 50 & 66.8  \\
    & CR-Net~\cite{Wang_IROS21_Cascaded} & & 33.2 & 27.8 & 34.3 & 47.2 & 60.8 \\
    & SGPA~\cite{chen_ICCV21_sgpa}&  &  37.1 & 35.9 & 39.6 & 61.3 & 70.7 \\ 
    & SPD~\cite{Tian_ECCV20_DeformNet} &  & 27.0 & 19.3 & 21.4 & 43.2 & 54.1 \\
    \cline{2-8}
    & SPD$^*$ & & 27.0 & 19.1 & 21.2 & 43.5 & 54.0 \\
    & SPD$^*$+Ours & \checkmark  & \textbf{43.6} &  \textbf{45.8} & \textbf{54.4} & \textbf{61.4} &\textbf{73.1} \\
    \hline
    \multirow{7}{*}{CAMERA25} & NOCS~\cite{wang2019nocs}&  & 37.0 & 32.3 & 40.9 & 48.2 & 64.6 \\
    & DualPoseNet~\cite{Lin_ICCV21_DualPoseNet} & \checkmark & 71.7 & 64.7 & 70.7 & 77.2 & 84.7\\
    & CR-Net~\cite{Wang_IROS21_Cascaded} &  & 75.0 & 72.0 & 76.4 & 81.0 & 87.7 \\
    & SGPA~\cite{chen_ICCV21_sgpa}&   & 69.1 & 70.7 & 74.5 & 82.7 & 88.4 \\
    & SPD~\cite{Tian_ECCV20_DeformNet} & & 46.9 & 54.3 & 59.0 & 73.3 & 81.5 \\
    \cline{2-8}
    & SPD$^*$ & & 46.9 & 54.1 & 58.8 & 73.9 & 82.1 \\
    & SPD$^*$+Ours & \checkmark & \textbf{76.1} & \textbf{75.4} & \textbf{80.3} & \textbf{83.3} & \textbf{89.3} \\
    \shline
\end{tabular}
}
\end{table*}

The quantitative results for category-level pose refinement on REAL275 and CAMERA25 are presented in \tbl{tab:cat_ref_sota}.
We use the results of SPD~\cite{Tian_ECCV20_DeformNet} as initial poses during inference in our main experiments and ablation studies.
It clearly shows that our method surpasses the baseline by a large margin, especially on the strict $\ang{5}\,\CM{2}$ metric, achieving an absolute improvement of 26.7\%  on the challenging REAL275 dataset and 21.3\% on the CAMERA25 dataset.

The results also demonstrate the significant superiority of CATRE over state-of-the-art methods~\cite{wang2019nocs,Lin_ICCV21_DualPoseNet,Wang_IROS21_Cascaded,chen_ICCV21_sgpa,Tian_ECCV20_DeformNet}. 
Notice that although we do not exploit any color information in the refinement procedure, we are still very competitive overall.
More importantly, on the strict metrics \iou{75} and $\ang{5}\,\CM{2}$, we surpass all the previous methods.
We kindly refer readers to the \emph{supplementary material} for more qualitative analyses and detailed results for each category.

\subsubsection{Ablation Study}

\begin{table*}[t]
\centering
\caption{
    {Ablation on different initial poses on REAL275.}
    \emph{(a)} The average precision (AP) of rotation ($^{\circ}$) and translation (cm) at different thresholds before and after refinement.
    \emph{(b)} Quantitative results. $^*$ denotes our re-implementation.
}
\begin{subfigure}[b]{0.45\linewidth}
    \includegraphics[width=\linewidth]{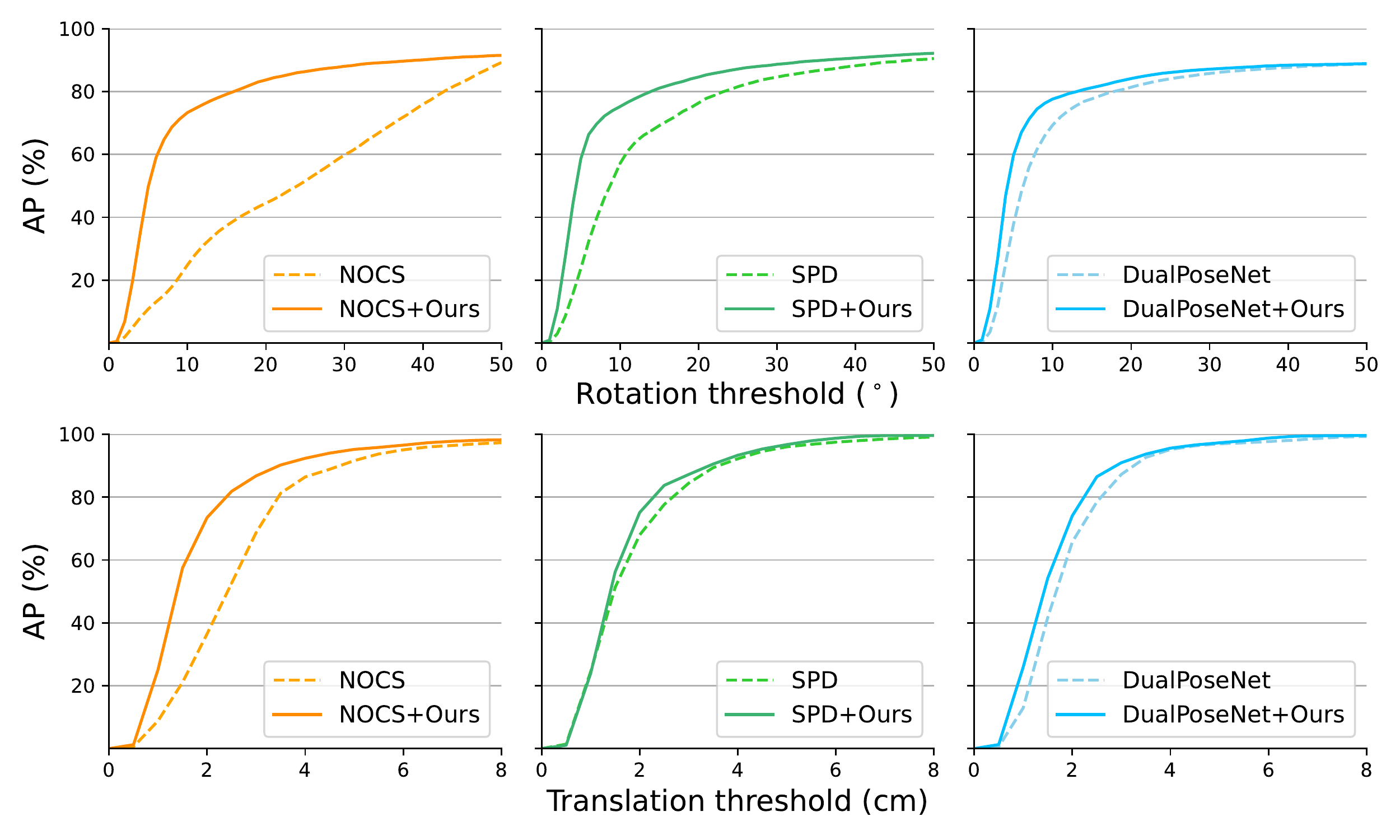}
    \caption{\label{fig:mAP_3methods}}
\end{subfigure}
\begin{subfigure}[b]{0.4\linewidth}
\scalebox{0.83}{
\tablestyle{2pt}{1.35}
\begin{tabular}{@{}l|c|c|cc@{}}
    \shline
    Method & Ref. &  \iou{75} & $\ang{5}\,\CM{2}$ & $\ang{5}\,\CM{5}$  \\
    \hline
    NOCS$^*$~\cite{wang2019nocs} &  & 9.0 & 7.3 & 9.9 \\
    w/ Ours & \checkmark & 42.6 & 40.7\textcolor{lime}{\tiny $\uparrow$\textbf{33.4}} & 47.8\textcolor{lime}{\tiny $\uparrow$\textbf{37.9}}  \\ 
    \hline
     SPD$^*$~\cite{Tian_ECCV20_DeformNet} &  & 27.0 & 19.1 & 21.2 \\
     w/ Ours & \checkmark & 43.6 &  45.8\textcolor{lime}{\tiny $\uparrow$\textbf{26.7}} & 54.4\textcolor{lime}{\tiny $\uparrow$\textbf{33.2}} \\
    \hline
     DualPoseNet$^*$\,\cite{Lin_ICCV21_DualPoseNet} & \checkmark &31.4 & 29.3 & 35.9\\
     w/ Ours & \checkmark & 44.4 & 43.9\textcolor{lime}{\tiny $\uparrow$\textbf{14.6}} & 54.8\textcolor{lime}{\tiny $\uparrow$\textbf{18.9}}\\ 
    \shline
\end{tabular}
}
\caption{\label{tab:sub_cat_initpose}}
\end{subfigure}\hfill
\label{tab:cat_initpose}
\end{table*}
\tbl{tab:cat_initpose} and \tbl{tab:all_cat_ref_abla} present the ablations on various factors.

\paragraph{Ablation on Different Initial Poses} 
To demonstrate the generalizability and robustness of our method \wrt different kinds of initial poses, we initialize the network with predictions of various qualities from~\cite{wang2019nocs,Tian_ECCV20_DeformNet,Lin_ICCV21_DualPoseNet} and present the results in \tbl{tab:cat_initpose}.
The network achieves consistent enhancement using the same weight,
no matter initialized with the early proposed methods \cite{wang2019nocs,Tian_ECCV20_DeformNet}, or recent state of the art \cite{Lin_ICCV21_DualPoseNet}.
\tbl{fig:mAP_3methods} further shows distinct improvement on rotation estimate after refinement.

\begin{table*}[t]
\centering
\caption{
    {Ablation studies on REAL275.}
    \emph{(a)} Accuracy and speed \wrt iterations.
    \emph{(b)} Quantitative results, where SPD$^*$ denotes our re-implementation of SPD~\cite{Tian_ECCV20_DeformNet}.
}
\begin{subfigure}[b]{0.26\linewidth}
    \includegraphics[width=\linewidth]{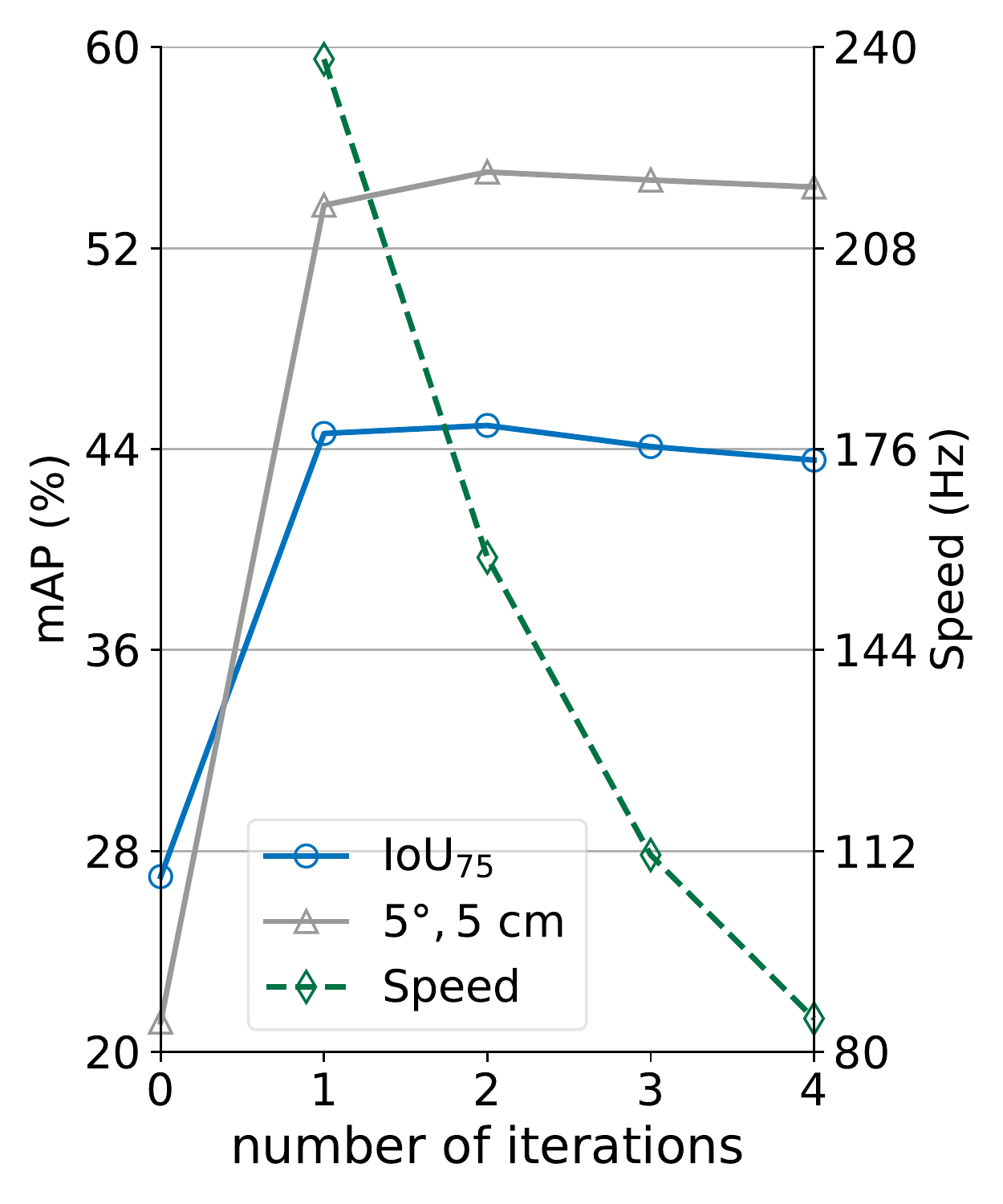}
    \caption{\label{fig:time_iter}}
\end{subfigure}
\begin{subfigure}[b]{0.7\linewidth}
\scalebox{0.7}{
\tablestyle{3pt}{1.2}
\begin{tabular}{@{}c|c|c|cc|c|c@{}}
    \shline
     Row & Method  &  \iou{75} & $\ang{5}\,\CM{2}$ & $\ang{5}\,\CM{5}$ & $\ang{5}$ & $\CM{2}$  \\
    \hline
    A0 & SPD$^*$ & 27.0 & 19.1 & 21.2 & 23.8 & 68.6 \\
    \hline
    B0 & SPD$^*$+Ours & 43.6 &  \textbf{45.8} &\textbf{54.4} & 58.0 & 75.1  \\
    \hline
    C0 & B0: Mean shape $\rightarrow$ 3D bounding box & 42.9 & 39.5 & 47.2 & 52.7 & 73.1  \\
    C1 & B0: Mean shape $\rightarrow$ 3D axes & 42.2 & 36.7 & 43.1 & 49.6 & 73.2 \\
    \hline
    D0 & B0: TS-Head also w/ $\text{MaxPool}(F_p)$ & \textbf{47.8} & 45.7 & 52.8 & \textbf{58.4} & 75.7  \\
    D1 & B0: w/o focalizing to origin & 41.4 & 40.2 & 47.5 & 52.3 & 70.7 \\
    D2 & B0: w/o adaptive points sampling & 29.0 & 26.7 & 42.3 & 46.3 & 63.3 \\
    \hline
    E0 & B0: PointNet\,\cite{qi2017pointnet} $\rightarrow$ 3D-GCN\,\cite{3DGC} & 28.4 & 36.0 & 43.4 & 47.7 & 68.0 \\
    E1 & B0: w/o weight sharing & 23.6 & 1.7 & 2.0 & 2.5 & 76.0 \\
    \hline
    F0 & B0: Single pose head (TS-Head style)& 46.4 & 40.7 & 46.2 & 52.6 & 75.9  \\
    F1 & B0: Single pose head (Rot-Head style) & 45.0 & 41.1 & 50.0 & 55.1 & 69.1 \\
    F2 & B0: Rot-Head conv fusion $\rightarrow$ MaxPool & 45.1 & 41.8 & 49.5 & 55.0 & 76.4 \\
    F3 & B0: Rot-Head conv fusion $\rightarrow$ AvgPool & 44.9 & 41.8 & 48.0 & 53.4 & \textbf{77.1} \\
    F4 & B0: TS-Head global feature$\rightarrow$point-wise & 29.8 & 32.3 & 46.3 & 53.0 & 57.5 \\
    \hline
    G0 & B0: w/o $\mathcal{L}_{pm}$ & 44.4 & 42.5 & 50.7 & 55.5 & 74.1  \\
    G1 & B0: w/o $\mathcal{L}_{\textbf{R}}$/$\mathcal{L}_{\textbf{t}}$/$\mathcal{L}_{\textbf{s}}$ & 38.4 & 22.6 & 30.1 & 34.0 & 72.8 \\
    \shline
\end{tabular}
}
\caption{\label{tab:cat_ref_abla}}
\end{subfigure}\hfill
\label{tab:all_cat_ref_abla}
\end{table*}

\paragraph{Ablation on Shape Prior}
Aside from the mean shape, CATRE can use the eight corners of the 3D bounding box ($N_p=8$) or four points constructing 3D axes ($N_p=4$) as shape prior,
the sizes of which are determined by $\size_{est}$.
As shown in \tbl{tab:cat_ref_abla}, the results using mean shape (B0) are superior to those using bounding box corners (C0) or axes (C1). 
However, the poses can still be refined taking bounding box corners or axes as input,
which implies that the geometric encoder is able to encode the initial pose information from sparse input without categorical prior knowledge.
Inspired by this, we further extend our method to pose refinement on the unseen category in Sec.~\ref{sec:cross-ref}.

\paragraph{Effectiveness of Point Cloud Focalization}
To verify the efficiency and accuracy of predicting relative pose from the focalized point clouds, we conduct several experiments. 
Firstly, the input of TS-Head is replaced with a concatenated feature from both $\text{MaxPool}(F_o)$ and $\text{MaxPool}(F_p)$. Since the relative translation can be derived from $\hat{\mathcal{O}}$ due to the focalization operation, adding feature from $\hat{\mathcal{P}}$ introduces unnecessary noise thus leading to a slight performance drop (\tbl{tab:cat_ref_abla} B0 \vs~D0).
Moreover, we translate the center of shape prior $\shp$ to $\trans_{init}$, and keep the pose of the observed point cloud $\pcl$ as it is in camera space. The initial translation is then appended to the input of TS-Head for better predicting $\trans_\Delta$. 
However, we observe a larger decrease in accuracy (\tbl{tab:cat_ref_abla} B0 \vs~D1) discarding focalization,
thus limiting the absolute range of input point clouds is vital for reducing the complexity of learning relative transformation.
To avoid introducing too many background points, we employed an adaptive strategy to sample points within a ball determined by the initial pose estimate.
Without this step, the accuracy decreased dramatically (\tbl{tab:cat_ref_abla} B0 \vs~D2).

\paragraph{Effectiveness of Shared Encoder}
As mentioned in \sect{sec:key_ingredients}, 3D-GCN~\cite{3DGC} performs worse than PointNet~\cite{qi2017pointnet} in our work (\tbl{tab:cat_ref_abla} B0 \vs~E0).
Moreover, weight sharing in the encoder is prominent for capturing relevant features regarding relative transformation by embedding $\hat{\mathcal{P}}$ and $\hat{\mathcal{O}}$ into a unified space, which is especially crucial to the prediction of $\rot_{\Delta}$.
Otherwise, it is almost unable to perform any refinement on rotation, as shown in \tbl{tab:cat_ref_abla} (B0 \vs~E1).

\paragraph{Effectiveness of Disentangled Heads}
We consider two types of single pose heads, \ie, extending TS-Head to estimate $[\mathbf{r}_{\Delta}^x|\mathbf{r}_{\Delta}^y]$ (F0), and adding two FC layers in Rot-Head to predict $\trans_{\Delta}$ and $\size_{\Delta}$ (F1).
\tbl{tab:cat_ref_abla} reveals that the disentangled design (B0) outperforms a single pose head leveraging a global feature (F0) or point-wise features (F1) consistently.

\paragraph{The Design Choices of Heads}
When replacing the trainable point-wise features aggregation with non-parametric operation MaxPool or AvgPool in Rot-Head, the accuracy on $\ang{5}$ drops signiﬁcantly (\tbl{tab:cat_ref_abla} B0 \vs~F2, F3).
Furthermore, treating $\trans$ and $\size$ as global attributes rather than collecting point-wise features like Rot-Head also enhances performance (\tbl{tab:cat_ref_abla} B0 \vs~F4).

\paragraph{Effectiveness of Loss Function}
\tbl{tab:cat_ref_abla} (B0 \vs~G0) evinces that $\loss_{pm}$ coupling pose prediction brings performance enhancement, and the individual loss components designated for $[\rot_{est}|\trans_{est}|\size_{est}]$ are crucial as well (\tbl{tab:cat_ref_abla} B0 \vs~G1).

\paragraph{Accuracy and Speed vs. Iteration}
In \tbl{fig:time_iter}, we show the accuracy  and inference speed \wrt iterations.
Here we empirically set 4 as the maximum number of iteration to balance the performance and speed.
On a machine with a TITAN XP GPU and two Intel 2.0GHz CPUs, CATRE runs at $\approx$85.32Hz 
for 4 iterations.

\subsection{Category-level Pose Tracking}

\begin{table*}[t!]
\tablestyle{3pt}{1.1}
\centering
\caption[caption]{Results of category-level pose tracking on REAL275. Following \cite{weng2021captra}, we perturb the ground-truth pose of the first frame as the initial pose.
}
\label{tab:tracking}
\scalebox{0.95}{
\begin{tabular}{@{}l|ccccc@{}}
    \shline
    Method & Oracle ICP\,\cite{weng2021captra} & 6-PACK\,\cite{Wang_ICRA20_6PACK} & iCaps\,\cite{deng2022icaps} & CAPTRA\,\cite{weng2021captra} & Ours \\
    \hline 
    Init.  & GT. & GT. Pert. & Det. and Seg. & GT. Pert. &  GT. Pert. \\
    \hline
    Speed (Hz) $\uparrow$ & - & 10 & 1.84 & 12.66 & \textbf{89.21} \\
    \hline
    mIoU $\uparrow$ & 14.7 & - & - & 64.1 & \textbf{80.2} \\
    \rete $\uparrow$ & 0.7 & 33.3 & 31.6 & \textbf{62.2} & 57.2 \\
    $\rot_{err} (^\circ) \downarrow$ & 40.3 & 16.0& 9.5 & \textbf{5.9} & 6.8 \\
    $\trans_{err} (\CM{}) \downarrow$ & 7.7 & 3.5 & 2.3 & 7.9 & \textbf{1.2}\\
    \shline
\end{tabular}
 }
\end{table*} 
\tbl{tab:tracking} summarizes the quantitative results for category-level pose tracking, with a comparison between our methods and other tracking-based methods~\cite{Wang_ICRA20_6PACK,deng2022icaps,weng2021captra}.

Leveraging a single model for tracking all objects, our method achieves competitive results with previous state-of-the-art methods.
Noticeably, the pipeline can run at 89.21Hz, which is $7\x$ faster than the state-of-the-art method CAPTRA~\cite{weng2021captra} and qualified for real-time application.
Moreover, our method exceeds CAPTRA in the metric of mIoU and $\trans_{err}$ by a large margin, in that CAPTRA solves $\trans$ and $\size$ using the indirect Umeyama algorithm without RANSAC, making their method sensitive to outliers.
In contrast, our network predicts $\trans$ and $\size$ in an end-to-end manner.
Additionally, our method deals with all the objects by a single network, while \cite{weng2021captra} needs to train separately for $\rot$ and $\trans$/$\size$ of each category.
To sum up, although CAPTRA performs slightly better than ours \wrt \rete~and $\rot_{err}$, 
we still have an edge over them in tracking speed and translation accuracy, which are vital in many robotic tasks such as grasping.

\subsection{Instance-level Pose Refinement}

\begin{table*}[t]
\caption{\label{tab:inst_ref_sota}
Results on LM referring to the Average Recall (\%) of ADD(-S) metric. $^*$ denotes symmetric objects. Ours(B) and Ours(F) use 8 bounding box corners and 128 FPS model points as shape priors respectively.
}
\centering
\scalebox{0.8}{
\begin{threeparttable}[c]
\tablestyle{1pt}{1.3}
\begin{tabular}{@{}lc ccccccccccccc c@{}}
\shline
Method & Ref. &  Ape & Bvise & Cam & Can & Cat & Drill & Duck  & Ebox$^*$ & Glue$^*$  & Holep & Iron  & Lamp  & Phone & Mean\\
\hline
SSD-6D~\cite{kehl2017ssd}+ICP &  \checkmark & 65 & 80 & 78 & 86 & 70 & 73 & 66 & \textbf{100} & \textbf{100} & 49 & 78 & 73 & 79 & 79 \\
DenseFusion~\cite{wang2019densefusion} & \checkmark & 92.3 & 93.2 & 94.4 & 93.1 & 96.5 & 87.0 & 92.3 & 99.8 & \textbf{100} & 92.1 & 97 & 95.3 & 92.8 & 94.3 \\
CloudAAE~\cite{gao2021cloudaae}+ICP & \checkmark & - & - & - & - & - & - & - & - & - & - & - & - & - & 95.5 \\
\hline
PoseCNN~\cite{posecnn} & & 27.9 & 69.4 & 47.7 & 71.1 & 56.4 & 65.4 & 43.2 & 98.2 & 95.0 & 50.7 & 65.9 & 70.4 & 54.4 & 62.7 \\
w/ Ours(B) & \checkmark & 63.7 & 98.6 & 89.7 & 96.1 & 84.3 & 98.6 & 63.9 & 99.8 & 99.4 & 93.2 & 98.4 & 98.7 & 97.5 & 90.9 \\
w/ Ours(F) & \checkmark & \textbf{94.1} & \textbf{99.5} & \textbf{97.9} & \textbf{99.4} & \textbf{98.0} & \textbf{99.8} & \textbf{92.9} & 99.8 & 99.7 & \textbf{98.3} & \textbf{99.0} & \textbf{99.8} & \textbf{99.5} & \textbf{98.3} \\
\shline
\end{tabular}
\end{threeparttable}
}
\end{table*}
Different from category-level refinement, instance-level pose refinement assumes the CAD model is available.
Considering this, we propose two types of shape priors, \ie, the tight bounding box of CAD model or $N_p$ model points selected by farthest point sampling (FPS) algorithm, where we choose $N_p=128$ empirically.

\tbl{tab:inst_ref_sota} shows the experimental results on LM dataset.
Our methods achieve significant improvement against the initial prediction provided by PoseCNN~\cite{posecnn}.
Noteworthy, Ours(B) utilizing the sparse bounding box corners information can still be comparable with methods refined by ICP procedure leveraging exact CAD models.
Furthermore, by explicitly exploiting the shape information with 128 FPS model points, Ours(F) achieves state of the art distinctly surpassing previous RGB-D based methods \cite{kehl2017ssd,wang2019densefusion,gao2021cloudaae} with refinement.

\subsection{Pose Refinement on Unseen Category}
\label{sec:cross-ref}

\begin{table*}[t]
\centering
\caption{
    {Results of unseen category pose refinement on YCB-Video.}
    \emph{(a)} Qualitative results, where white, red and green contours illustrate the ground-truth, initial (modified from PoseCNN~\cite{posecnn}) and refined (Ours) poses, respectively.
    \emph{(b)} Quantitative results (Metric: AUC of ADD(-S)).
}
\begin{subfigure}[b]{0.4\linewidth}
    \centering
    \includegraphics[width=0.7\linewidth]{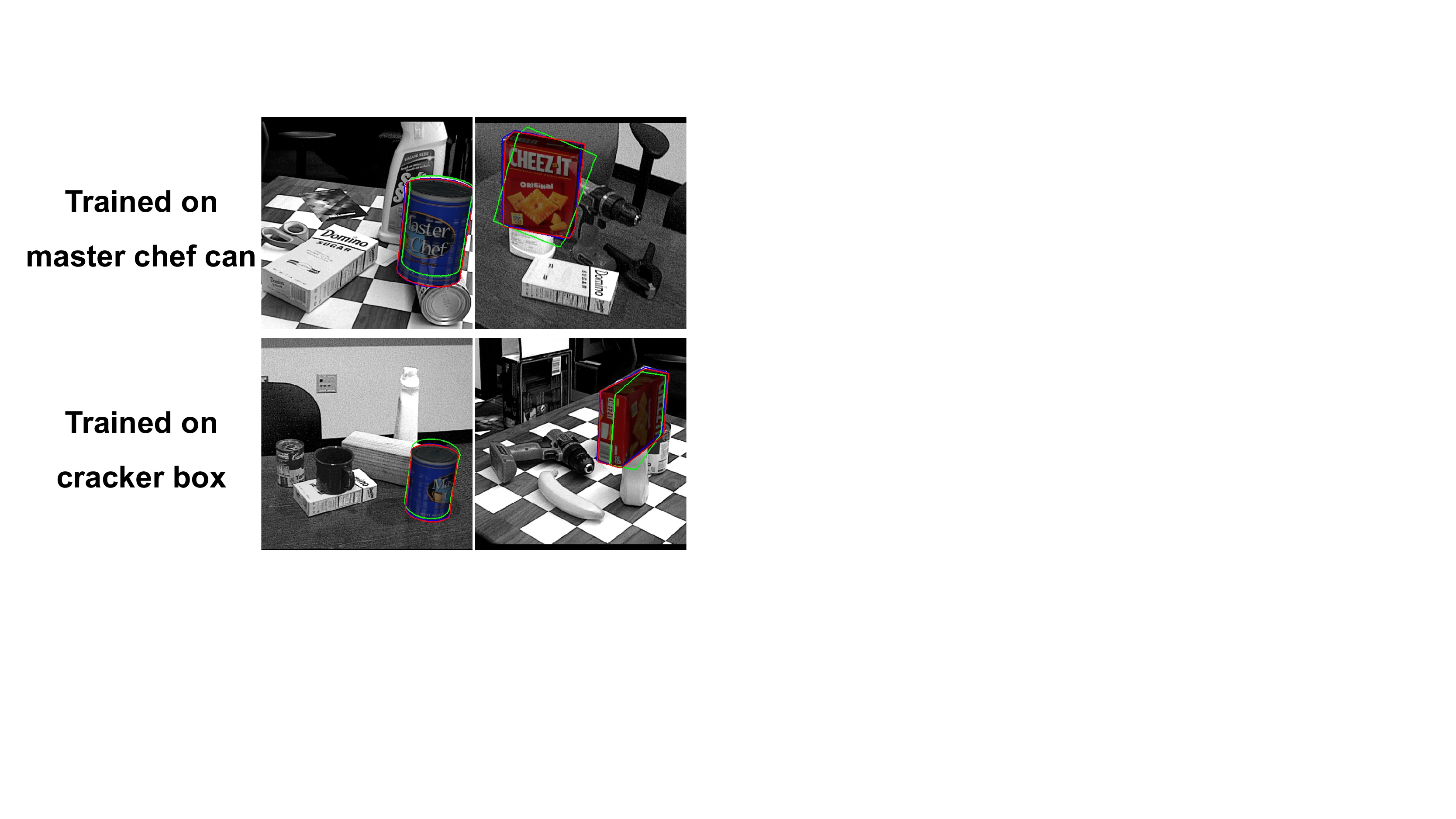}
    \caption{\label{fig:ycbv_transfer}}
\end{subfigure}~~
\begin{subfigure}[b]{0.4\linewidth}
\centering
\tablestyle{3pt}{1.1}
\begin{tabular}{@{}l|cc|cc@{}}
    \shline
    \multirow{2}{*}{\diagbox{Train}{Test}} & \multicolumn{2}{c|}{Master Chef Can} & \multicolumn{2}{c}{Cracker Box} \\
    \cline{2-5}
    & \textcolor{red}{Init.} & \textcolor{green}{Ref.} & \textcolor{red}{Init.} & \textcolor{green}{Ref.} \\
    \hline
    Master Chef Can & 52.7 & 71.3 & 57.8 & 65.3\textcolor{lime}{\tiny $\uparrow$\textbf{7.5}} \\
    Cracker Box & 52.7 & 57.4\textcolor{lime}{\tiny $\uparrow$\textbf{4.7}} & 57.8 & 89.6 \\
    \shline
\end{tabular}
\caption{\label{tab:sub_ycbv_transfer}}
\end{subfigure}\hfill
\label{tab:ycbv_transfer}
\end{table*}

As discussed before, utilizing bounding box corners as shape prior offers a possibility for pose refinement on unseen category, as long as the objects have a unified definition of the canonical view.
We verify this assumption in two instances on the challenging YCB-V dataset, \ie, master chef can and cracker box, and present qualitative and quantitative results in \tbl{tab:ycbv_transfer}.
Training with master chef can, the network is able to refine the pose of the unseen category cracker box and vice versa.
It demonstrates that the network learns some general features for pose refinement
exceeding the limitation of the trained category.
We hope this experiment will inspire future work for unseen object refinement or tracking leveraging self-supervision or domain adaptation techniques.
\section{Conclusion}
We have presented CATRE, a versatile approach for pose refinement and tracking, which enables tackling instance-level, category-level and unseen category problems 
with a unified framework.
The key idea is to iteratively align the focalized shape prior and observed point cloud by predicting a relative pose transformation using disentangled pose heads.
Our algorithm is generalizable to various kinds of baselines and achieves significant boosts in performance over other state-of-the-art category-level pose estimation works as well as competitive results with tracking-based methods.
Future work will focus on more challenging scenarios, such as 
tracking unseen objects, leveraging RGB information, and training with only synthetic data utilizing large-scale 3D model sets~\cite{chang2015shapenet,collins2021abo,reizenstein21co3d}.

\noindent \textbf{Acknowledgments}
We thank Yansong Tang at Tsinghua-Berkeley Shenzhen Institute, Ruida Zhang and Haotian Xu at Tsinghua University for their helpful suggestions.
This work was supported by the National Key R\&D Program of China under Grant 2018AAA0102801 and National Natural Science Foundation of China under Grant 61620106005.

%
%
\bibliographystyle{splncs04}
\bibliography{ref_long_short_names,ref}

\clearpage

\renewcommand\thefigure{\thesection.\arabic{figure}}
\renewcommand\thetable{\thesection.\arabic{table}}
\setcounter{figure}{0} 
\setcounter{table}{0} 

\appendix

\section*{Appendix}

\section{More Implementation Details}

\subsection{Architecture of Encoder}

\begin{figure}[h]
\begin{center}
  \includegraphics[width=\linewidth]{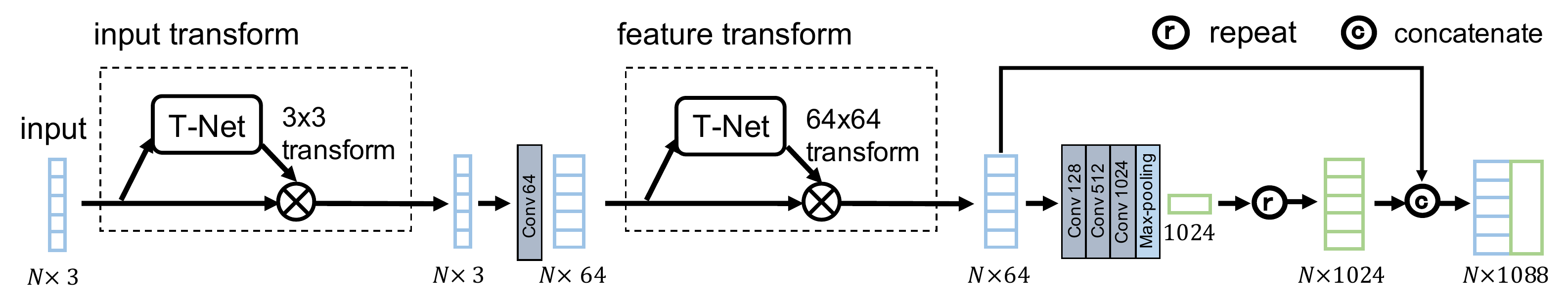}
\end{center}
  \caption{
  Architecture of the PointNet-based~\cite{qi2017pointnet} geometry encoder, where $N$ denotes the number of input points.
  }
\label{fig:backbone}
\end{figure}

\noindent \fig{fig:backbone} depicts the detailed architecture of our employed PointNet-based~\cite{qi2017pointnet} geometry encoder as mentioned in Sec. 3.3 of the main text.
In specific, the encoder first maps each point from $\mathbb{R}^3$ into a high-dimensional space, and generates an additional global feature by aggregating features from all points. 
The global feature is then repeated $N$ times and concatenated to each point-wise feature, where $N$ denotes the number of input points.

\begin{figure}[t]
\begin{center}
  \includegraphics[width=\linewidth]{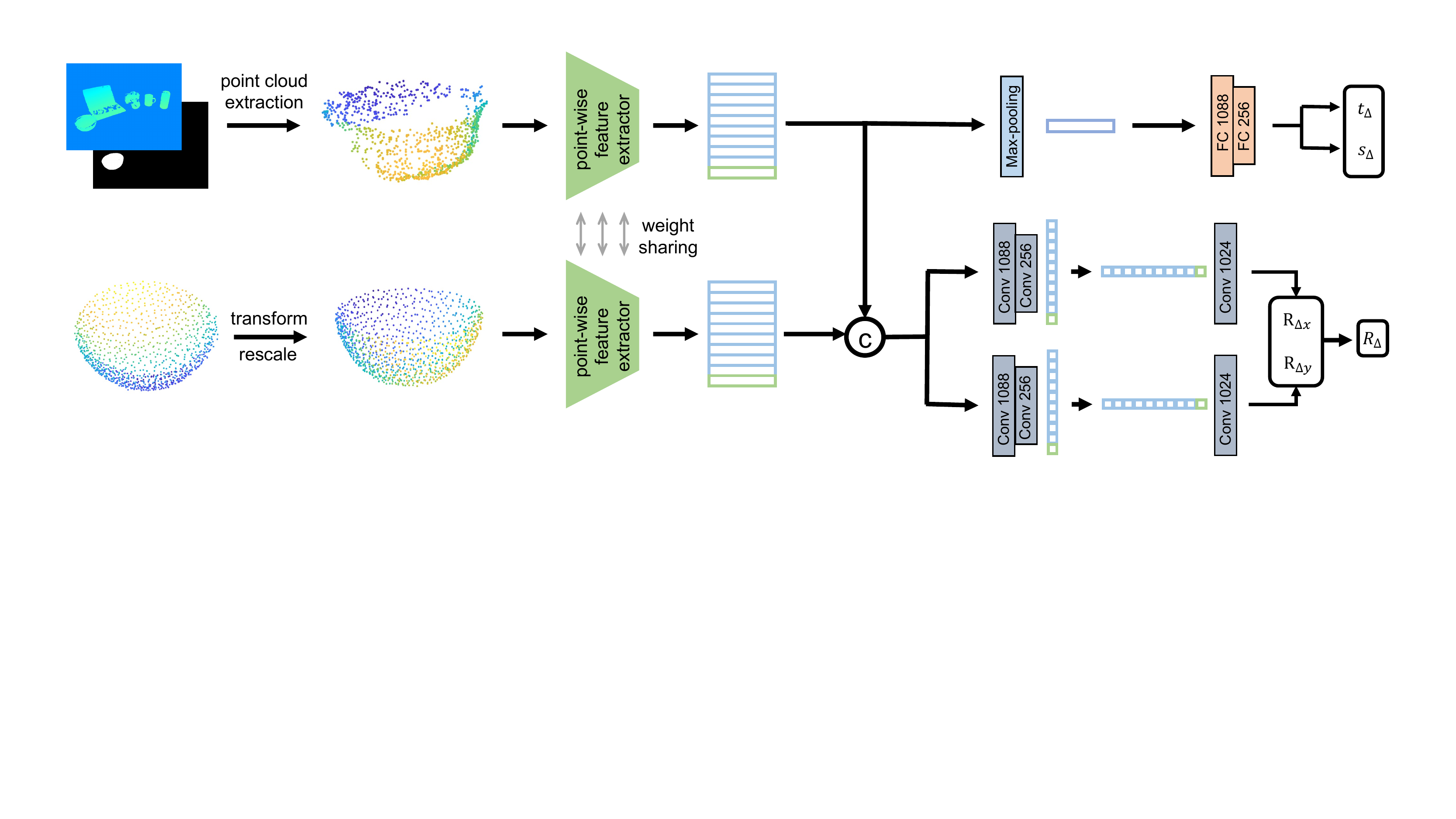}
\end{center}
  \caption{
  Network architecture of the instance-level refiner, notice that $\size_{init}$ and $\size_{\Delta}$ related components are removed compared to the category-level refiner CATRE.}
\label{fig:instre}
\end{figure}

\subsection{Instance-level Pose Refinement}
As shown in \fig{fig:instre}, since the instance-level refiner (see Sec. 4.4 of the main text) only estimates 6DoF pose transformation, we make some modification based on our CATRE network.
Specifically, the initial size information $\size_{init}$ is removed in both the focalization operation and the original TS-Head,
and the output FC layer for relative size prediction is also discarded, making TS-Head into T-Head.
Accordingly, the loss function is changed to
\begin{equation}
    \mathcal{L} = \mathcal{L}_{pm} + \mathcal{L}_{\rot} + \mathcal{L}_{\trans},
\end{equation}
thereby
\begin{equation}
\begin{cases}
    \mathcal{L}_{pm} &= \underset{\textbf{x} \in \mathcal{P}}{\avg}\|(\rot_{gt}\textbf{x} + \trans_{gt}) - (\rot_{est}\textbf{x} + \trans_{est})\|_1, \\ 
    \mathcal{L}_{\rot} &= \frac{3-\Tr(\rot_{gt}\rot_{est}^T)}{4}, \\
    \mathcal{L}_{\trans} &= \|\trans_{gt} - \trans_{est} \|_1.
\end{cases}
\end{equation}

\section{Additional Results}
\subsection{Accuracy on NOCS \vs Iterations}
\begin{table*}[t!]
\tablestyle{18pt}{1.1}
\centering
\caption[caption]{Accuracy \wrt iterations. We initialize the predictions with the reproduced results from NOCS~\cite{wang2019nocs}.
}
\label{tab:iter_nocs}
\begin{tabular}{@{}l|ccccc@{}}
    \shline
    Iteration & 0 & 1 & 2 & 3 & 4  \\
    \hline
    \iou{75}          & 9.0 & 21.8 & 30.2 & 39.9 & \textbf{42.6} \\
    $\ang{5}\,\CM{2}$ & 7.3  & 29.0 & 34.2 & 39.1 & \textbf{40.7} \\
    \rete             & 9.9  & 30.7 & 36.9 & 44.2 & \textbf{47.8} \\
    $\ang{5}$         & 10.8 & 31.9 & 38.6 & 46.1 & \textbf{49.7} \\
    $\CM{2}$          & 37.8 & 71.6 & 74.5 & 74.2 & \textbf{74.8} \\
    \shline
\end{tabular}
\end{table*} 
Apart from the results in Sec.~4.2 of our paper, in \tbl{tab:iter_nocs}, we show the additional results for accuracy \wrt iterations, where the initial poses are provided by our reproduced results from NOCS~\cite{wang2019nocs}.
Since the initial predictions from NOCS are poor, more iterations of refinement improve the results substantially.
Considering the trade-off between performance and speed, we empirically set 4 as the maximum number of iteration in all the refine experiments.

\subsection{Category-level Refinements on Each Category}
\begin{table*}[t!]
\tablestyle{1pt}{1.1}
\centering
\caption[]{Detailed results of category-level pose refinement for each category on REAL275. $^*$  denote our re-implementation of the method.}
\label{tab:ref_each}
\tablestyle{3pt}{1.0}
\begin{tabular}{@{}l|c|cc|cc|cc@{}}
    \shline
    \multicolumn{2}{c|}{Method} & NOCS$^*$ &  w/ Ours & SPD$^*$ & w/ Ours & DualPoseNet$^*$  &  w/ Ours \\
    \hline
    \multirow{3}{*}{bottle} & \iou{75}           & 8.8 & 25.9 & 12.2 & 24.5 & 12.7 & 16.4 \\
                            & $\ang{5}\,\CM{2}$  & 2.6 & 42.9 & 21.6 & 57.0 & 27.8 &43.1 \\
                            & \rete              & 2.6 & 49.2 & 23.2 & 61.6 &33.4 &56.7\\
    \hline 
    \multirow{3}{*}{bowl} & \iou{75}             & 42.8 & 88.0 & 76.1 & 88.2 & 80.8 & 92.6\\
                          & $\ang{5}\,\CM{2}$    & 40.2 & 84.5 & 50.5 & 87.1 & 72.0 &91.4\\
                          & \rete                & 49.8 & 85.8 & 54.0 & 91.1 &78.8 &97.4\\
    \hline 
    \multirow{3}{*}{camera} & \iou{75}           & 0.8 & 4.1 & 3.4 & 5.9 & 1.4 & 4.1\\
                          & $\ang{5}\,\CM{2}$    & 0.0 & 0.5 & 0.0 &0.8 & 0.0& 0.1 \\
                          & \rete                & 0.0  & 0.6 & 0.0 & 0.9 &0.0 &0.1\\
    \hline 
    \multirow{3}{*}{can}  & \iou{75}             & 1.5 & 14.4 & 29.6 & 33.5 & 5.3 & 16.2\\
                          & $\ang{5}\,\CM{2}$    & 0.3 & 62.4 & 37.9 & 66.0 & 32.3 &49.1\\
                          & \rete                & 0.3 & 70.7 & 42.7 & 79.3 &47.2 &67.4\\
    \hline 
    \multirow{3}{*}{laptop} & \iou{75}           & 0.4 & 78.7 & 32.7 & 71.6 & 66.3 & 71.3\\
                          & $\ang{5}\,\CM{2}$    & 0.6 & 39.5 & 4.6 & 51.8 & 36.7 &56.8\\
                          & \rete                & 6.5 & 65.8 & 7.0 & 81.1& 48.8& 83.8\\
    \hline 
    \multirow{3}{*}{mug} & \iou{75}              & 0.0 & 44.3 & 8.0 & 37.6 & 21.6 & 65.9\\
                          & $\ang{5}\,\CM{2}$    & 0.0 & 14.4 & 0.3 & 11.8 & 7.0& 23.0\\
                          & \rete                & 0.0 & 14.7 & 0.3 & 12.6 & 7.1 &23.5 \\
    \hline 
    \multirow{3}{*}{overall}& \iou{75}           & 9.0 & 42.6 & 27.0 & 43.6 & 31.4 & 44.4\\
                            & $\ang{5}\,\CM{2}$  & 7.3 & 40.7 & 19.1 & 45.8 & 29.3 &43.9\\
                            & \rete              & 9.9 & 47.8 & 21.2 & 54.4 &35.9 &54.8\\
    \shline
\end{tabular}
\end{table*} 

\tbl{tab:ref_each} evinces category-level pose refinement results for each category.
Notably, our method achieves consistent enhancement on 6 categories \wrt each metric.
However, the refined pose for camera is still poor and the reasons are twofold.
First, the initial pose prediction is far from ground truth, making the refinement procedure extremely hard.
Moreover, the shape variance among the camera category is huge, so that the categorical mean shape is not certainly reliable.

Since CATRE is meant for real-time applications, it is vital whether the model is able to scale with different sizes and resolutions.
The resolution and size have minimal impact on the inference speed because we uniformly sample a fixed number of points (\ie, 1024) from the input point cloud.
This strategy is applied vastly \cite{g2lnet,Tian_ECCV20_DeformNet,weng2021captra,Chen_CVPR21_FSNet} and scales well with CATRE.
Concretely, there is about ten times the difference between the resolution of ``laptop'' and ``bottle'', but CATRE can refine their poses with one model (\tbl{tab:ref_each}).

\subsection{Qualitative Results of Category-level Pose Refinement on REAL275}
\begin{figure}[tb!]
\begin{center}
  \includegraphics[width=0.95\linewidth]{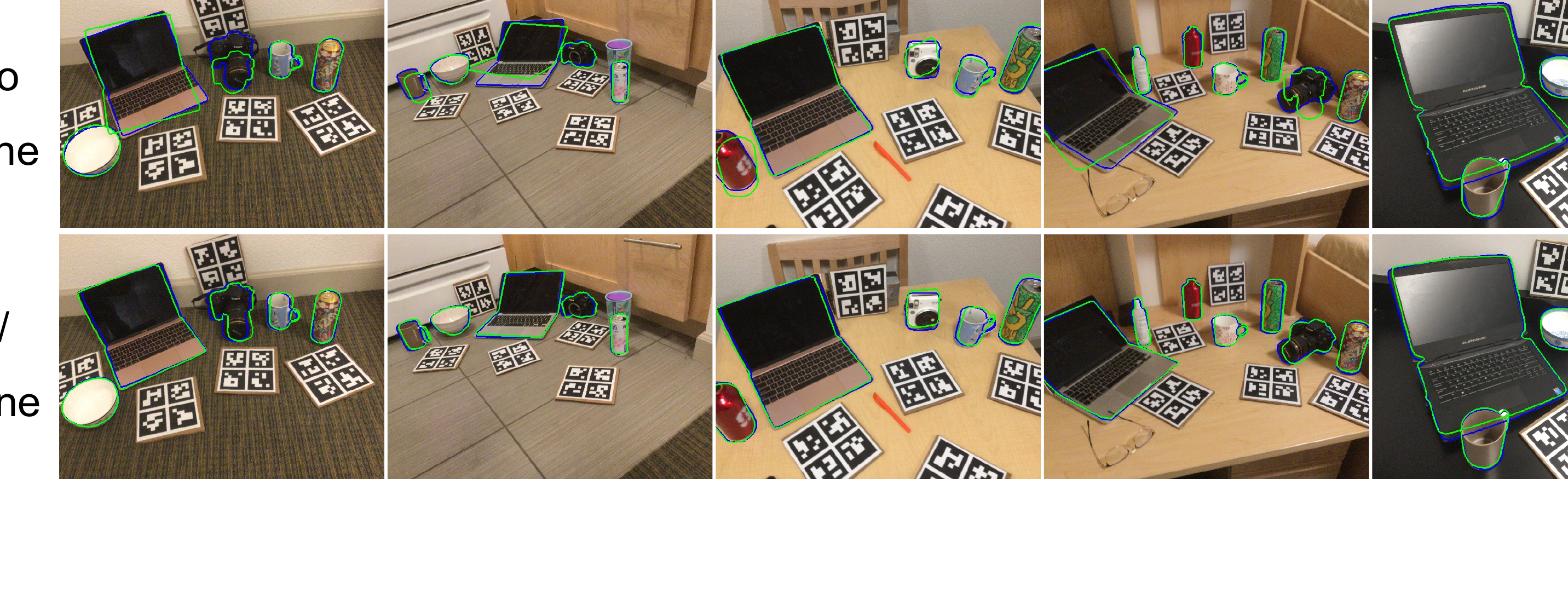}
\end{center}
  \caption{Qualitative comparison, where white, red and green contours demonstrate ground-truth, initial (SPD~\cite{Tian_ECCV20_DeformNet}) and refined (Ours) poses, respectively. The axis length reflects predicted scale meanwhile.}
\label{fig:refine}
\end{figure}

\fig{fig:refine} presents the qualitative examples of category-level pose refinement on REAL275. 

\subsection{More Ablation Studies on REAL275}

\begin{table*}[t]
\centering
\caption{
    {Additional ablation studies on REAL275.}
    \emph{(a)} Accuracy \wrt additional noises, where $\times m$ means adding $m$ times of training noise to grund-truth poses during inference.
    \emph{(b)} Quantitative results, where SPD$^*$ denotes our re-implementation of SPD~\cite{Tian_ECCV20_DeformNet}. 
}
\begin{subfigure}[b]{0.26\linewidth}
    \includegraphics[width=\linewidth]{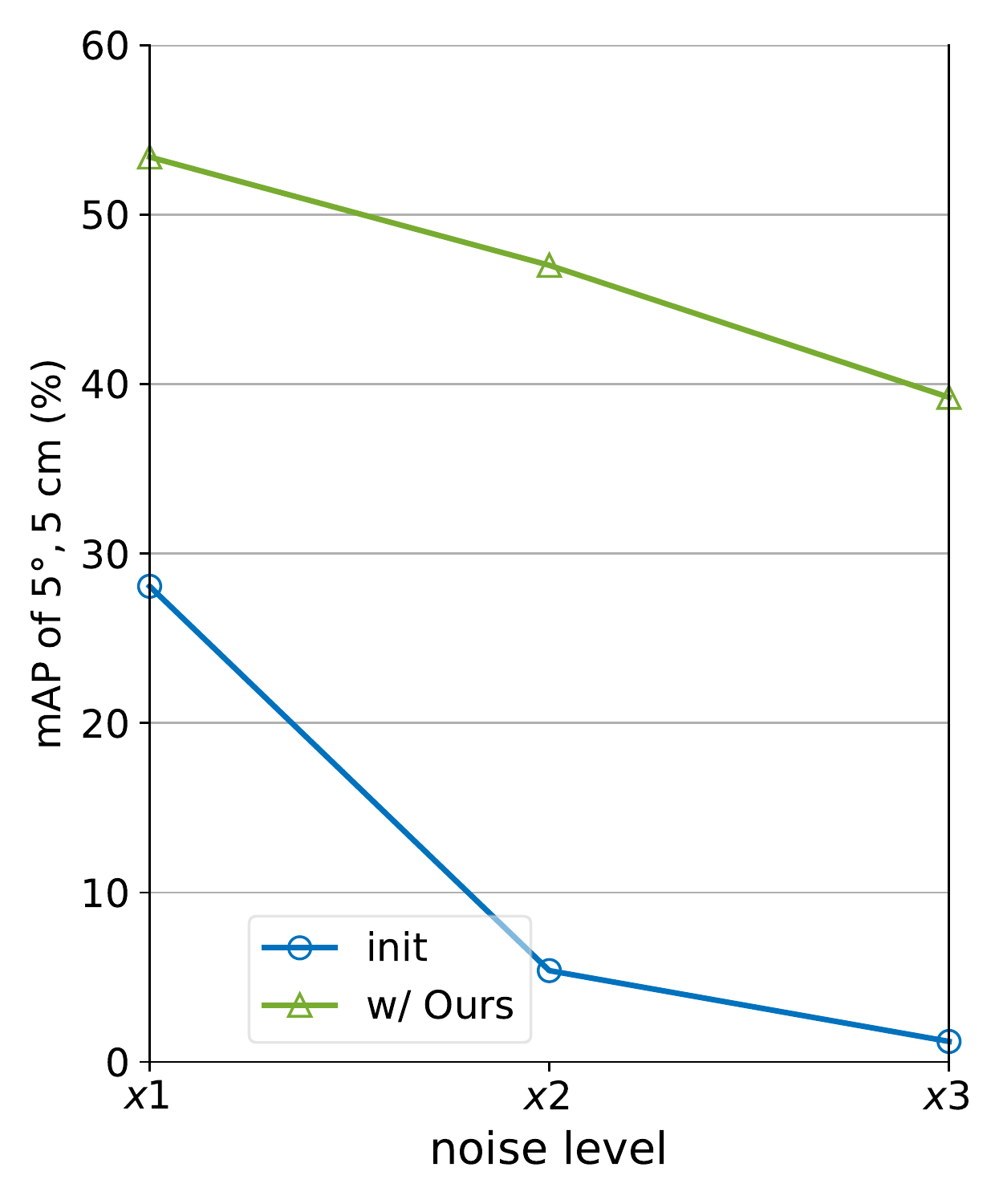}
    \caption{\label{fig:gt_noise}}
\end{subfigure}
\begin{subfigure}[b]{0.7\linewidth}
\scalebox{0.7}{
\tablestyle{8pt}{2.0}
\begin{tabular}{@{}c|c|c|cc|c|c@{}}
    \shline
     Row & Method  &  \iou{75} & $\ang{5}\,\CM{2}$ & $\ang{5}\,\CM{5}$ & $\ang{5}$ & $\CM{2}$  \\
    \hline
    A0 & SPD$^*$ & 27.0 & 19.1 & 21.2 & 23.8 & 68.6 \\
    \hline
    B0 & SPD$^*$+Ours & 43.6 &  \textbf{45.8} &\textbf{54.4} & \textbf{58.0 }&\textbf{ 75.1 } \\
    \hline
    C0 & B0:  w/o T-Net in PointNet & \textbf{43.7} & 41.8 & 50.0 & 55.3 & 74.1 \\
    C1 & B0:  w/  occlusion & 36.5 & 33.4 & 38.6 & 43.8 & 67.3 \\
    \shline
\end{tabular}
}
\caption{\label{tab:cat_more_abla}}
\end{subfigure}
\label{tab:cat_more_ablations}
\end{table*}

More ablation studies on category-level pose refinement are presented in \tbl{tab:cat_more_ablations}.

We conduct refinement on the initial poses obtained by perturbing the ground truth with Gaussian noise to explore the sensitivity of CATRE \wrt large noise.
As shown in \tbl{fig:gt_noise},
thanks to the iterative testing strategy, even with harsh noise existing in initial poses, CATRE can still achieve satisfying results.

\tbl{tab:cat_more_abla} shows some additional results. 
By removing T-Net in PointNet-based \cite{qi2017pointnet} geometry encoder, the accuracy \wrt rotation and translation decreases.
Since rare occlusion exists in REAL275 dataset, we cropped a 25\% block around a random corner of predicted bounding boxes to imitate occlusion.
We find that CATRE can exceed the baseline (\tbl{tab:cat_more_abla} C1 \vs A0), but occlusion still has a side effect on performance (\tbl{tab:cat_more_abla} C1 \vs B0).

\subsection{BOP Results on LM}

\begin{table*}[t]
\caption{\label{tab:bop_metrics}
Results on LM referring to the Average Recall (\%) of BOP metric. $^*$ denotes symmetric objects. Ours(B) and Ours(F) use 8 bounding box corners and 128 FPS model points as shape priors respectively.
}
\centering
\begin{threeparttable}[c]
\tablestyle{10pt}{1.1}
\begin{tabular}{@{}l ccc c@{}}
\shline
Method  &  $\text{AR}_\text{{\tiny MSPD}}$ & $\text{AR}_\text{{\tiny MSSD}}$ & $\text{AR}_\text{{\tiny VSD}}$ & AR\\
\hline
PoseCNN~\cite{posecnn} & 86.1 & 73.5 & 59.3 & 73.0 \\
w/ Ours(B)            & 74.7 & 88.4 & 91.4 & 84.8 \\
w/ Ours(F)            & 84.2 & 95.1 & 96.4 & 91.9 \\
\shline
\end{tabular}
\end{threeparttable}
\end{table*}

In the main text, we evaluate our method on LM by traditional protocol following the compared methods~\cite{wang2019densefusion,gao2021cloudaae,kehl2017ssd}.
Recently, BOP metric~\cite{hodan2020bop} proves to be more suited for the evaluation of pose estimation, specifically for symmetric objects.
It report an average recall (AR) by calculating the mean score of three metrics: $\text{AR}=(\text{AR}_\text{{\tiny MSPD}} + \text{AR}_\text{{\tiny MSSD}} + \text{AR}_\text{{\tiny VSD}})/3$. 
Details of these metircs can be referred to \cite{hodan2020bop}.

As shown in \tbl{tab:bop_metrics}, our method still achieve distinct enhancement \wrt AR against the initial prediction provided by PoseCNN~\cite{posecnn}.

\subsection{Failure Cases Analyses}

\begin{figure}[h!]
\begin{center}
  \includegraphics[width=\linewidth]{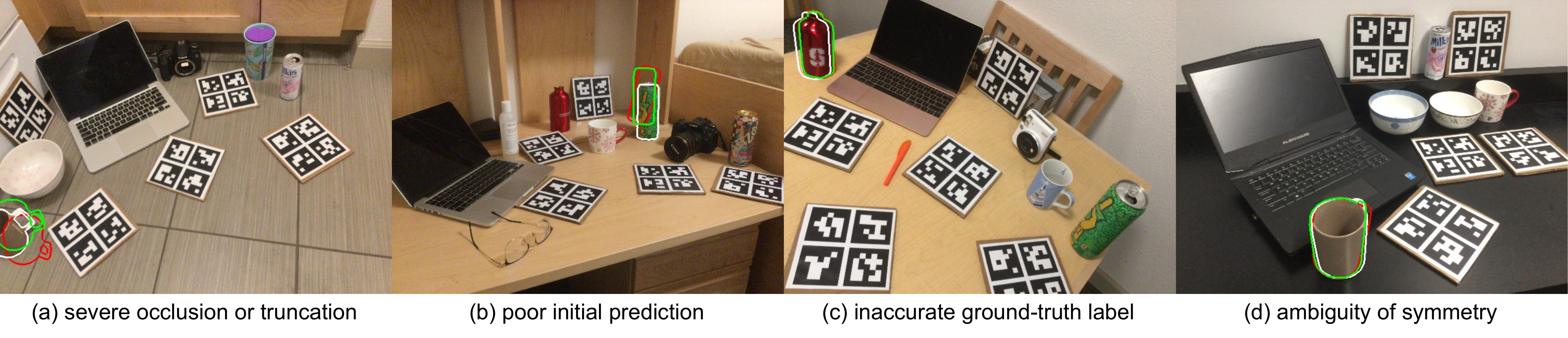}
\end{center}
  \caption{
   Several failure cases,  where white, red and green contours demonstrate ground-truth, initial (SPD~\cite{Tian_ECCV20_DeformNet}) and refined (Ours) poses, respectively.}
\label{fig:failure_cases}
\end{figure}

As shown in \fig{fig:failure_cases}, failure conditions can be generally grouped into 4 categories:
(a) severe occlusion or truncation, (b) poor initial prediction, (c) inaccurate ground-truth label, and (d) ambiguity of symmetry (especially for mug and camera).

\end{document}